\documentclass[letterpaper]{article} 
\usepackage{aaai24}  
\usepackage{times}  
\usepackage{helvet}  
\usepackage{courier}  
\usepackage[hyphens]{url}  
\usepackage{graphicx} 
\usepackage{natbib}  
\usepackage{caption} 
\usepackage{algorithm}
\usepackage{algorithmic}
\usepackage{amsmath}

\usepackage{newfloat}
\usepackage{listings}
\DeclareCaptionStyle{ruled}{labelfont=normalfont,labelsep=colon,strut=off} 
\floatstyle{ruled}
\newfloat{listing}{tb}{lst}{}
\floatname{listing}{Listing}
\usepackage{xspace}
\usepackage{enumitem}
\usepackage{subfigure}
\usepackage{multirow}
\usepackage{booktabs}
\usepackage{amsfonts}
\usepackage{amsmath}
\newcommand{\method}{{FISEdit}\xspace}

\title{Accelerating Text-to-Image Editing via Cache-Enabled Sparse Diffusion Inference}
\author{
    Zihao Yu\textsuperscript{\rm 1},
    Haoyang Li\textsuperscript{\rm 1},
    Fangcheng Fu\textsuperscript{\rm 1},
    Xupeng Miao\textsuperscript{\rm 2},
    Bin Cui\textsuperscript{\rm 1,3} \\
}
\affiliations{
    \textsuperscript{\rm 1} School of Computer Science \& Key Lab of High Confidence Software Technologies (MOE), Peking University \\
    \textsuperscript{\rm 2} Carnegie Mellon University\\
    \textsuperscript{\rm 3} Institute of Computational Social Science, Peking University (Qingdao), China\\
    \{zihao.yu, ccchengff, bin.cui\}@pku.edu.cn, lihaoyang@stu.pku.edu.cn, xupeng@cmu.edu
}

\usepackage{bibentry}

\begin{document}

\maketitle

\begin{abstract}
    Due to the recent success of diffusion models, text-to-image generation is becoming increasingly popular and achieves a wide range of applications. Among them, text-to-image editing, or continuous text-to-image generation, attracts lots of attention and can potentially improve the quality of generated images. It's common to see that users may want to slightly edit the generated image by making minor modifications to their input textual descriptions for several rounds of diffusion inference. However, such an image editing process suffers from the low inference efficiency of many existing diffusion models even using GPU accelerators.

    To solve this problem, we introduce Fast Image Semantically Edit (\method), a cached-enabled sparse diffusion model inference engine for efficient text-to-image editing. The key intuition behind our approach is to utilize the semantic mapping between the minor modifications on the input text and the affected regions on the output image. For each text editing step, \method can 1) automatically identify the affected image regions and 2) utilize the cached unchanged regions' feature map to accelerate the inference process. For the former, we measure the differences between cached and ad hoc feature maps given the modified textual description, extract the region with significant differences, and capture the affected region by masks.
    For the latter, we develop an efficient sparse diffusion inference engine that only computes the feature maps for the affected region while reusing the cached statistics for the rest of the image. 
    Finally, extensive empirical results show that \method can be $3.4\times$ and $4.4\times$ faster than existing methods on NVIDIA TITAN RTX and A100 GPUs respectively, and even generates more satisfactory images.
\end{abstract}

\section{Introduction}

The recent months have witnessed the astonishing improvement of diffusion models. Text-to-image generation is one of the most popular applications of diffusion models, which involves generating realistic images based on textual descriptions.
While diffusion models have been shown to generate high-quality images with good coverage of the data distribution, they require substantial computational power to generate high-quality images. 
Thus, it is an emerging research area to make the generation process of diffusion models more effective and efficient.

\begin{figure*}[!t]
  \centering
  \includegraphics[width=0.80\linewidth]{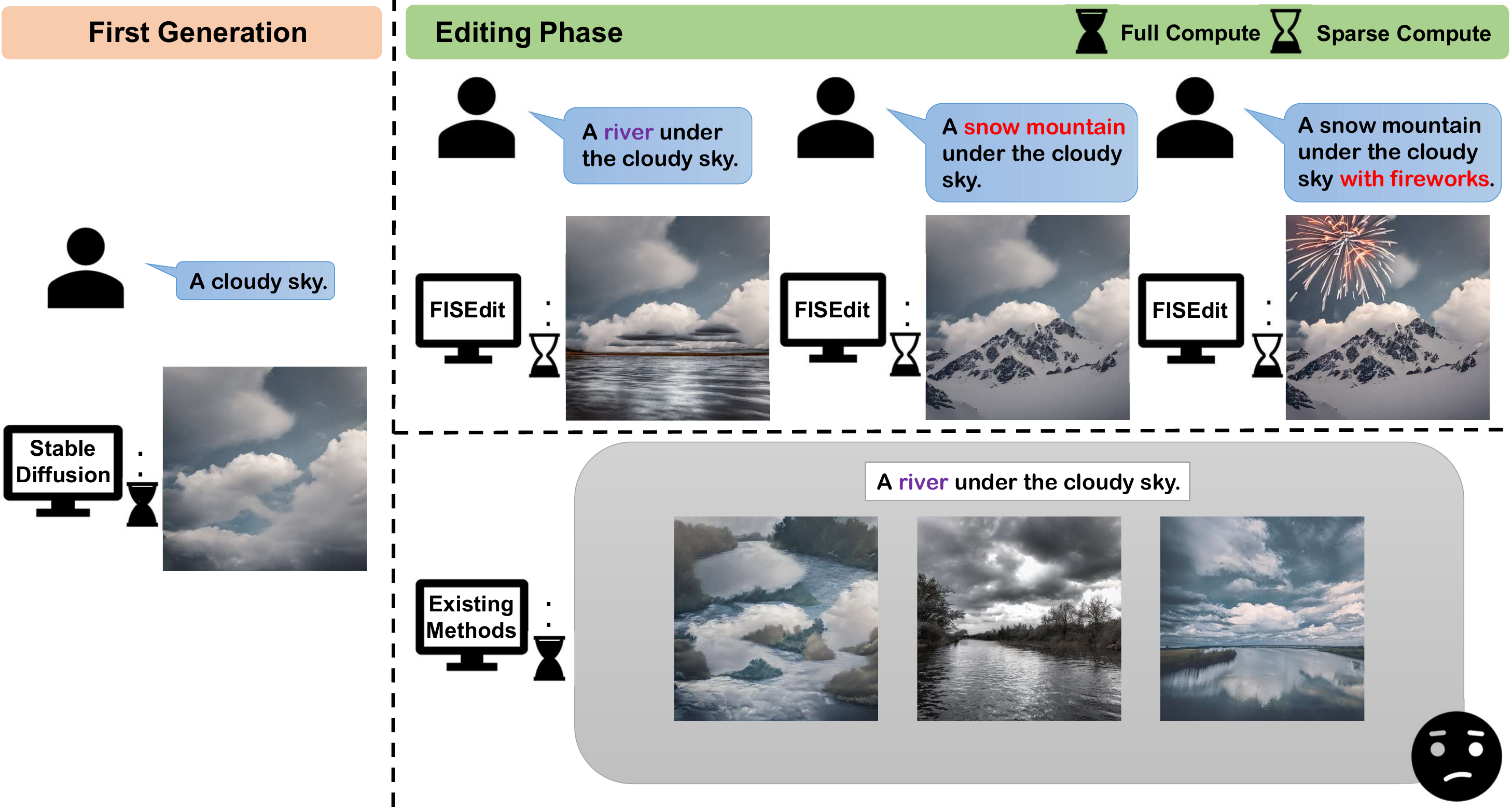}
  \caption{A real example of the user's interaction with \method and existing methods.}
  \label{samples}
\end{figure*}

\textbf{Our target scenario.} In real-world cases, after an image has been generated, it is common that the user may be unsatisfied with some specific regions of the generated image and desire to make subsequent \textbf{semantic edits}. Figure~\ref{samples} demonstrates a concrete example. The user may wish to interactively edit the image by making minor modifications to the textual description. These modifications involve adding specifications (e.g., adding a ``river'' or the ``fireworks'') or replacing words/phrases (e.g., replacing the ``river'' with ``snow mountain''). And in each round of editing, the user expects to make changes to only a small, specific region (affected region) and keep the rest (non-affected regions) untouched.

\textbf{Challenges.} However, we find that the current implementations fall short in supporting such a semantic editing task effectively and efficiently. First, due to the diversity nature of diffusion models, existing methods tend to make significant changes to the image. Although there are applications that allow users to provide masks to control the in-painting regions, it would lead to poor user experience. Second, even though only a small region is expected to be modified, existing works typically follow the vanilla inference process of diffusion models, which generates the entire image from scratch, without taking the spatial property into account. Undoubtedly, this is a waste of time and computational resources since the generation of non-affected regions is pointless.

In essence, to support the semantic editing task effectively and efficiently, there involves two technical issues: (1) \textit{how to automatically and accurately detect the affected region to be revised given the original image and the modified description} and (2) \textit{how to focus on the target region during the generation while skipping the non-affected regions?}

\textbf{Summary of contributions.} In this paper, we develop a brand new framework for the semantic editing task. 
The technical contributions of this work are summarized as follows.

\begin{itemize}[leftmargin=*]
\item We introduce Fast Image Semantically Edit (\method), a framework for minor image editing with semantic textual changes. \method is a direct application to the inference process of pre-trained diffusion models and does not need to train new models from scratch. To support the interactive semantic editing task, \method caches the feature maps of the latest generation process and addresses the two technical issues respectively.

\item We present a mask generation algorithm to capture the specific region for editing. To be specific, we quantify the distance between the cached and ad hoc latents given the modified textual description. Subsequently, we extract the region displaying substantial differences.

\item Based on the detected region and the cached feature maps, we propose a sparse updating approach to minimize unnecessary computational burden. More precisely, we develop an end-to-end sparse diffusion inference mechanism that selectively computes feature maps solely for the affected region and re-uses the cached statistics for the non-affected regions.

\item 
Compared with existing text-based image editing works, \method can generate competitive high-quality images that are consistent to the semantic edit. Furthermore, \method outperforms existing works in terms of generation efficiency by $4.4\times$ on NVIDIA TITAN RTX and $3.4\times$ on NVIDIA A100.
\end{itemize}

\section{Related Works}

Recent advances in diffusion models have enabled state-of-the-art image
synthesis~\cite{DDPM, DDIM, lu2022dpm, balaji2022ediffi, ramesh2022hierarchical, saharia2022photorealistic, huang2022fastdiff, bianchi2022easily, ramasinghe2020conditional}, including image editing and text-to-image generation. Compared to other generative models such as GANs~\cite{GAN2, GAN} and VAEs~\cite{VAE}, diffusion models have been shown to generate high-quality images with good coverage of the data distribution. However, they necessitate a substantial amount of computational power for both training and high-quality image generation, which makes them expensive and challenging to deploy in production systems.

\subsection{Image Editing Methods}

The user's requirement for image modification can be satisfied by \textbf{image editing models}~\cite{ zhu2016generative,abdal2019image2stylegan, abdal2020image2stylegan++, zhu2020sean, choi2021ilvr, kim2021diffusionclip, patashnik2021styleclip, rombach2022high, instructpix2pix, weictnet} and \textbf{controllable image generation methods}~\cite{SDEdit,
park2019semantic,
nichol2021glide,
hertz2022prompt,
couairon2022diffedit, orgad2023editing, parmar2023zero, feng2023near}.

\subsubsection{Image editing models} can accurately modify images, but they require to train new diffusion models, which is computationally expensive and time-consuming. Furthermore, most of the existing models are difficult to control the affected region without a user-drawn mask. We only compare our method with InstructPix2Pix~\cite{instructpix2pix} and stable diffusion in-painting models, since they have shown the best editing effect yet.

\subsubsection{Controllable image generation methods} are able to edit image using existing pre-trained weights, but they always require equal or even more computations to detect affected regions and meet textual change. For example, DIFFEdit~\cite{couairon2022diffedit} need extra reverse diffusion steps to calculate difference mask, and pix2pix-zero~\cite{parmar2023zero} fails to edit high-resolution images due to its overuse of memory.

\subsection{Diffusion Model Acceleration}
As diffusion model training and inference can be computationally expensive and time-consuming, especially when dealing with high-resolution images, various methods have been explored to accelerate the training and inference of diffusion models. Firstly, many recent works~\cite{DDPM, DDIM, lu2022dpm, lu2022dpmplus, watson2021learning, meng2022distillation, kong2021fast, xiao2021tackling, salimans2022progressive} have successfully reduced the number of iteration steps required for diffusion model inference while maintaining or improving the quality of the generated samples.
Moreover, some deep learning frameworks such as TensorFlow~\cite{abadi2016tensorflow} and PyTorch~\cite{paszke2019pytorch} speed up diffusion models by optimizing GPU kernels and paralleling computation streams.
All these research directions are orthogonal to ours and can be integrated with FISEdit.

Another way to accelerate diffusion models is reducing computation by exploiting the sparsity of the input data or the model parameters. 
There are several techniques~\cite{dong2017more, liu2018efficient, ren2018sbnet, liu2018dynamic, child2019generating, bolya2022token, han2023personalized, bi2023novel, zhangfine, yi2022flag} available for implementing sparse computation. 
However, directly applying existing techniques such as tiled sparse convolution~\cite{ren2018sbnet} fails to achieve considerable speedup and image quality in our scenario. 

In particular, recent works have successfully applied sparse computation to generative models~\cite{bolya2023token, sige}. SIGE~\cite{sige} is a general-purpose sparse inference framework for generative models, which integrates sparse convolution techniques into image editing tasks and makes sufficient use of data sparsity. However, SIGE~\cite{sige} faces limitations when applied to Stable Diffusion~\cite{rombach2022high} models due to its significant additional memory overhead requirement and inability to support textual changes. Therefore, we will not include SIGE as a baseline in the experimental part. In comparison to SIGE, our proposed framework supports more efficient sparse kernels and addresses the limitations encountered by SIGE, including batched input restrictions, intermediate activation management and fine-grained mask generation designed to detect areas affected by semantic changes.

\section{Method} 
\begin{figure*}[!t]
  \centering
  \includegraphics[width=0.8\linewidth]{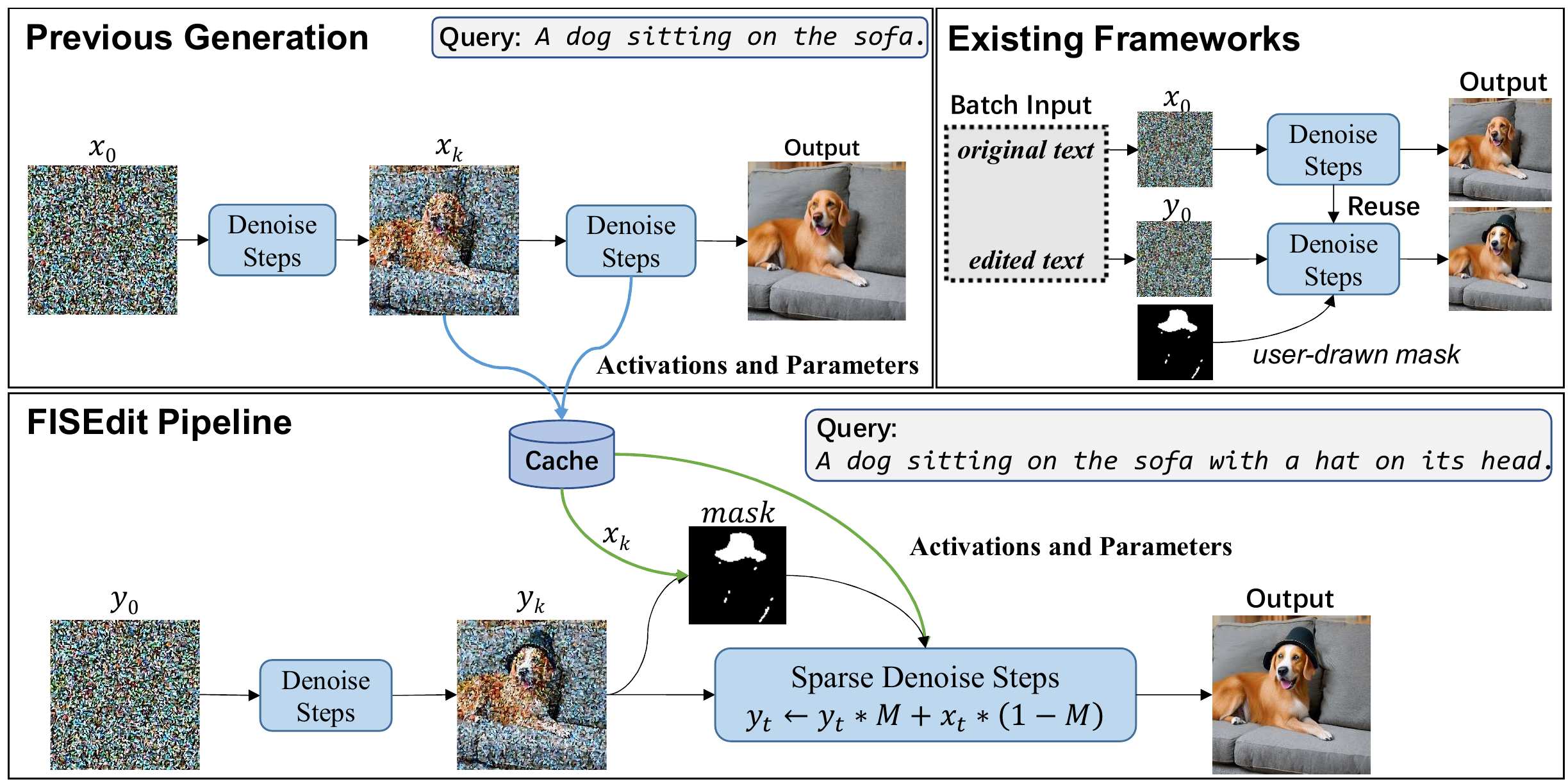}
  \caption{Overview structure of \method. When a query arrives, our system first executes $k$ denoise steps, and then generates a difference mask according to the output latents of $k$ steps. In the remaining denoise steps, the pre-computed results (activations and parameters) of each layers in U-Net will be reused according to the mask and the feature maps will be computed sparsely. Compared to existing frameworks which leverage the batched inputs, we collect and cache the results of previous generation to avoid redundant computation, and use the mask to control as well as accelerate the new T2I generation process at U-Net level.}
  \label{overview}
\end{figure*}

Figure~\ref{overview} shows the overview structure of \method. This method mainly derives from our observation that by sparsely calculating the crucial region of the feature maps in U-Net according to a fine-grained mask, we can effectively maintain the original image structure as well as get close to the new prompt semantics. Meanwhile, the sparsity of the computation could also benefit a lot to the model efficiency. Below, we first introduce a method to detect affected areas and generate difference mask for the subsequent sparse computation. Then we present our improved implementation of sparse computation algorithms, along with a non-trivial image editing pipeline.

\subsection{Fine-Grained Mask Generation}

Difference masks are used in image editing tasks to index changed regions, but existing models have sub-optimal performance when the user cannot draw a precise mask. To address this, we propose a mask generation method, which automatically identifies the image areas that require substantial change according to the modification of textual description. 

\begin{figure}[h]
  \centering
    \includegraphics[width=\linewidth]{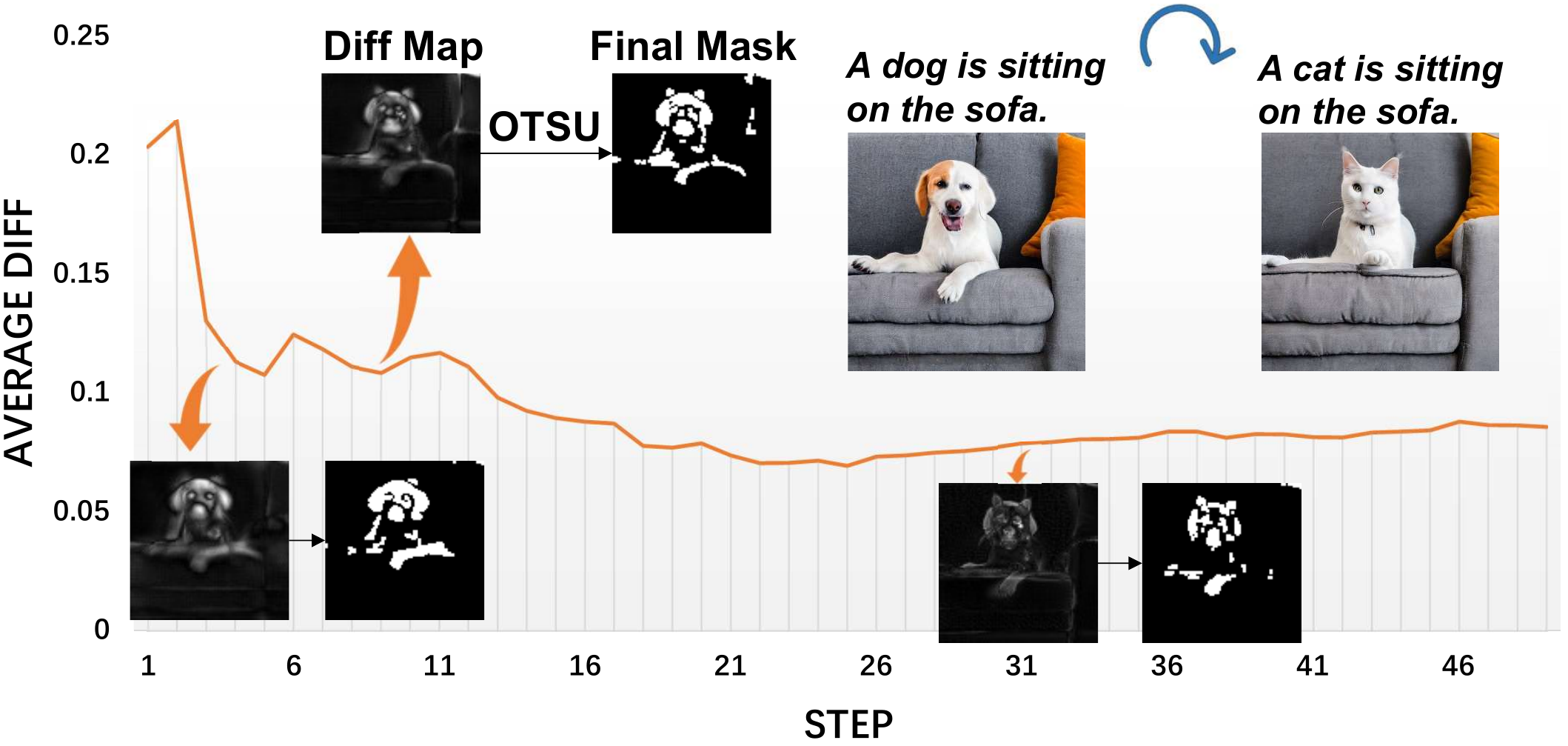}
  \caption{Variation of latent difference with iteration steps.}
  \label{diff_trend}
\end{figure}

\paragraph{Target Area Capture.}
To detect the affected areas in feature maps, we observe the feature maps in each step of the Stable Diffusion~\cite{rombach2022high} inference process and compare them using two similar textual prompts. Unfortunately, even though we have controlled the random factors, the changed regions remain large and unpredictable. Drawing inspiration from Prompt-to-Prompt~\cite{hertz2022prompt}, these irregular spatial changes can be controlled by enhancing cross-attention~\cite{crossattention} module. Specifically, we cache and preserve part of the cross-attention map corresponding to the unchanged text prompts in the calls with new description, and then the changed regions can be structured and easily located through few denoise steps. As demonstrated in Figure~\ref{diff_trend}, the difference caused by textual change can be observed clearly in step 5 to 10 and convert to a mask after finding the proper threshold $\epsilon$ via the following formula:
\begin{align*}
& diff :=~Normalize\left(\sum_{t=t_1}^{t_2}\lvert X_t-Y_t\rvert\right),~t_1,t_2\in [1,10] \\
& ~~~~~A := \{i|diff_{pixel_i}<\epsilon\}, B := \{j|diff_{pixel_j}\ge\epsilon\} \\
& ~~~~~~~~~~~~~~~~~N_1 :=Card(A), N_2 :=Card(B) \\
& \underset{\epsilon\in[0,1]}{\arg\max}~OTSU(\epsilon) :=  \frac{N_1N_2}{\left(N_1+N_2\right)^2}\bigg(\frac{1}{N_1}\sum_{i\in A}diff_{pixel_i} \\
& ~~~~~~~~~~~~~~~~~~~~~~~~~~~~~~~~~-\frac{1}{N_2}\sum_{j\in B}diff_{pixel_j}\bigg)^2
\end{align*}

Consequently, we no longer need extra diffusion steps to calculate the mask as our previous work did (e.g., DIFFEdit~\cite{couairon2022diffedit}).

\paragraph{Sparsity Analysis.}
\label{sparsity-analysis}

We evaluate our method by generating mask for the human-written dataset used by InstructPix2Pix~\cite{instructpix2pix}, where each sample in the dataset contains an original and manually modified caption for an image.

We find that the average edited size for these samples is less than 30\%, reflecting an astonishing fact that most image regions can be reused and numerous calculations can be potentially saved. However, when the feature map is compressed to low-resolution, sparsity decreases correspondingly. Consequently, we adopt the sparse computation for high-resolution inputs only, and forgo those with low-resolution.

\subsection{Sparse Computation}

In order to accelerate diffusion models and control the semantic changes, we focus on the U-Net~\cite{unet} modules since they are the most time-consuming and influential components in the image generation process.

The U-Net architecture is composed of several attention and ResNet~\cite{he2016deep} modules. Based on the observation in the previous section, it is practicable to perform sparse calculations on a small proportion of the whole image in the above modules. Below, sparse optimization on these modules will be discussed respectively.

\subsubsection{Adaptive Pixel-Wise Sparse Convolution.}
Tiled sparse convolution algorithm was proposed in SBNet~\cite{ren2018sbnet} to handle sparse data within ResNet modules. However, as we mentioned in Section~\ref{sparsity-analysis}, sparsity will decrease when processing low-resolution feature maps, and this is particularly obvious in tiled sparse convolution layers. Therefore, directly applying existing technique fails to achieve considerable speedup and image quality in our scenario.

Given that the efficiency of this algorithm is largely contingent on the distribution of difference masks and the size of split blocks, the constant block-gather strategy used in SBNet doesn't seem plausible. Therefore, it is essential to select a compatible block size with the given mask. We introduce Adaptive Pixel-wise Sparse Convolution (APSC) to ensure tiled sparse convolution achieves desired effect when dealing with capricious mask. We denote $(h,w)$, $N_{h,w}^{m}$ as the block's spatial size and the number of blocks gathers using given mask $m$ respectively, and assume the kernel size is $(h_k, w_k)$. We search for the optimal division strategy heuristically via the following approximation:
\[
\underset{h,w\in \{2,4,8,16,32\}}{\arg\min}f(h,w): =(h-h_k+1)*(w-w_k+1)*N_{h,w}^{m}
\]

All the variables mentioned above can be determined during the preprocessing stage, ensuring that APSC does not introduce any additional computational costs to model inference.

\subsubsection{Approximate Normalization.}
Since some normalization layers such as group normalization~\cite{wu2018group} entail calculating the mean and variance of the feature map in spatial dimension, it is infeasible to use sparse feature map in these layers. 
Inspired by batch normalization~\cite{ioffe2015batch}, which uses the mean and variance during training for model inference, it is plausible to use approximate methods in normalization layer. 
With the assumption that a tiny textual modification will not affect the distribution of the whole image, we are able to cache the mean and variance computed in the previous image generation and directly reuse them to facilitate the sparse inference process. Therefore, without the reliance on the statistics of the input feature map, sparse computation is available in all normalization layers.

\begin{figure}[!t]
  \centering
  \includegraphics[width=\linewidth]{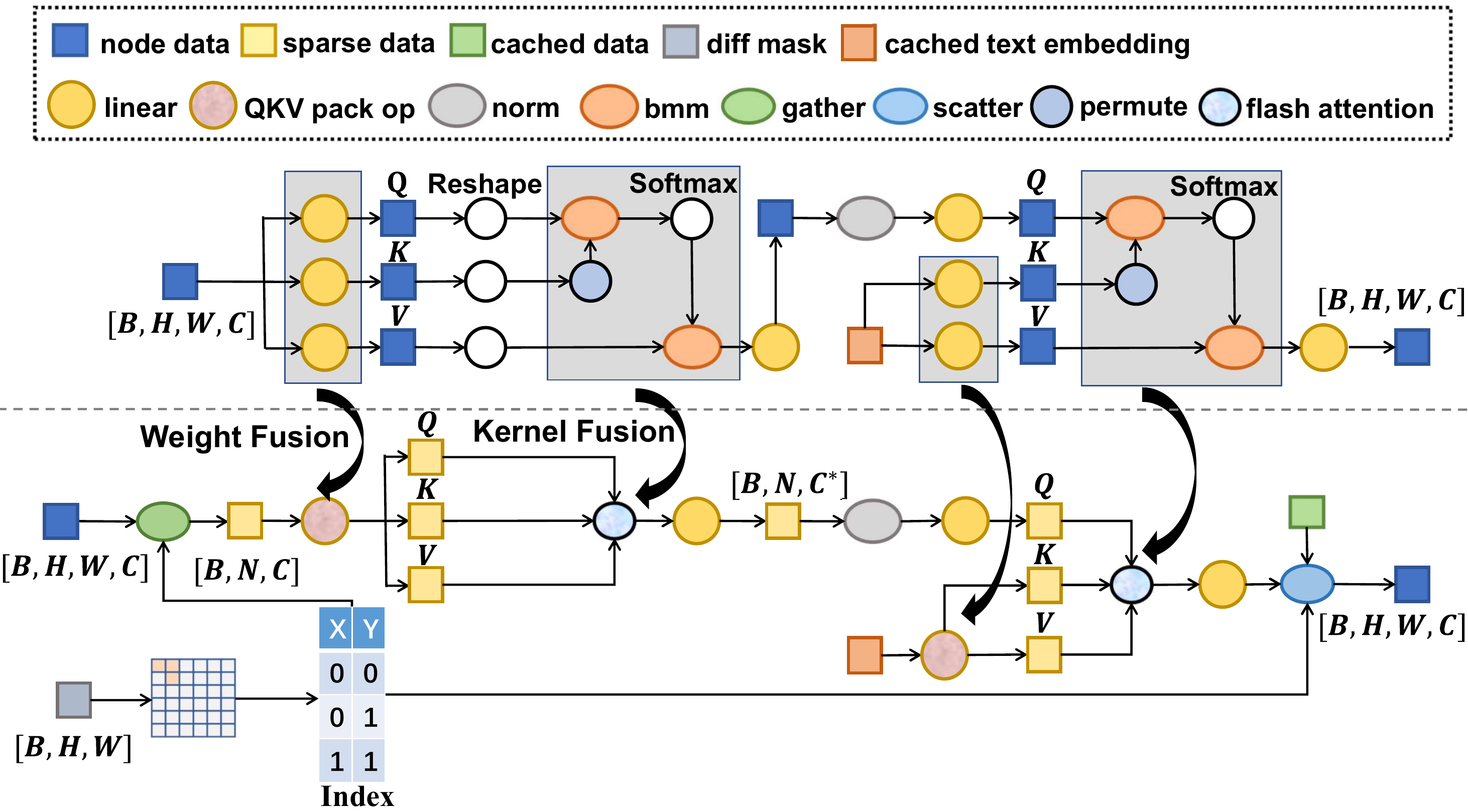}
  \caption{Overall improvement of the attention module.}
  \label{attention}
\end{figure}

\subsubsection{Approximate Attention.}

Recent diffusion models have incorporated self-attention~\cite{vaswani2017attention} and cross-attention~\cite{crossattention} modules to enhance the image quality and the correlation between image and text in text-to-image tasks, resulting in computational bottlenecks. Readers might suspect whether it is possible to utilize the sparsity of edited areas in attention layers, since the whole feature map is required to be used in self-attention layers in order to learn the long-term relevance among pixels. Intuitively, if we only calculate the results using non-zero regions in difference mask, the output will be inevitably deviated from the accurate result. Fortunately, such precision errors can be controlled to a tolerable range through some accuracy compensation measures. Specifically, we only apply sparse computation in the self-attention layers with a high-resolution input feature map. In these layers, non-zero regions account for only a small proportion of the entire feature map, and their influence on the unaffected area is consequently negligible, particularly after the SoftMax layer.

Figure~\ref{attention} shows the overview of our implementation of attention modules. Masked regions are collected before being input into the subsequent self-attention and cross-attention layers, utilizing sparse data exclusively within the attention layers to avoid unnecessary data movement.

\subsection{Cache-Based Editing Pipeline}

In order to support sparse computation, we have to store activation and parameters in the prior inference procedure, and load them in the following image generations. However, for the diffusion models, it's imperative to store and load activations for all iteration steps (e.g., 50 for DDIM~\cite{DDIM} scheduler and 10 for DPM-Solver++~\cite{lu2022dpmplus}), implying that there will be a significant amount of extra memory storage. For instance, when we edit an image with a $768\times768$ resolution, the memory storage for each iteration step is up to 4.2G, culminating in a total memory storage of up to 210 GB for 50 steps. Obviously, additional devices (e.g., CPU, HDD) are required for storage, and dealing with the movement of such a substantial amount of data during inference also presents a formidable challenge.

As depicted in Figure~\ref{overview}, existing image semantic editing methods operate under an inefficient working mode, which relies on batch input of the text before and after modification, in order to avoid the storage and loading of intermediate activation. Unfortunately, this mechanism involves a significant amount of repetitious computation. In fact, data movement and kernel computation are asynchronous on GPU, so the data transfer time can be overlapped by kernel computation. However, as our method dramatically reduces the computational time, data transmission will inevitably become a new bottleneck. To resolve this impasse, we introduce a cache-based editing pipeline to support asynchronous data movement and distributed storage.
Concretely, our system automatically moves stored data according to the memory usage of CPU and GPU, making the data load and storage completely asynchronous.
Besides, we reuse the memory of tensors with same shapes during the inference so that memory usage can be lowered.

\section{Experiments}
We implement our system based on the HuggingFace's diffusers\footnote{https://github.com/huggingface/diffusers}, which is a generic framework for training and inference of diffusion models. We clone this project and integrate it with our self-developed sparse inference engine Hetu\footnote{https://github.com/PKU-DAIR/Hetu/}~\cite{miao2021het,DBLP:journals/chinaf/MiaoXP22,miao2022hetgmp}.

\paragraph{Baselines.} We compare our method with the following baselines, and all the implementations for them can be found in HuggingFace's diffusers. All the baselines and our method use the pre-trained weights and configurations of the Stable Diffusion v2 model\footnote{https://huggingface.co/stabilityai/stable-diffusion-2}, and more details about our evaluation configurations can be found in our repository\footnote{https://github.com/Hankpipi/diffusers-hetu}.

\begin{itemize}[leftmargin=*]
\item \emph{Vanilla stable diffusion text-to-image pipeline (SDTP).} A simple attempt to edit image is to fix the internal randomness (e.g., the initial latent) and regenerate using the edited text prompt.
\item \emph{Stable diffusion in-painting pipeline (SDIP).} A variant of stable diffusion models, aiming at painting from a draft and text description. User-drawn masks are required in this model, and we share the difference mask generated by our method with it in order to make a fair comparison.

\item \emph{Prompt-to-Prompt}~\cite{hertz2022prompt}. A text-driven image editing method based on cross-attention control.

\item \emph{SDEdit}~\cite{SDEdit}. An image editing method  based on a diffusion model generative prior, which synthesizes realistic images by iteratively denoising through a stochastic differential equation.

\item \emph{InstructPix2Pix}~\cite{instructpix2pix}. A model that edits images according to user-written instructions instead of descriptions.

\item \emph{DIFFEdit}~\cite{couairon2022diffedit}. An image semantic editing method with automatic mask generation algorithm.

\item \emph{PPAP}~\cite{tumanyan2023plug}. A semantic editing framework based on self-attention optimizations.

\item \emph{Pix2Pix-Zero}~\cite{parmar2023zero}. An image-to-image translation method
that can preserve the content of the original image without manual prompting.

\end{itemize}

\paragraph{Datasets.}
We select LAION-Aesthetics~\cite{schuhmann2022laion} as the evaluation dataset, and follow the preprocessing method of InstructPix2Pix. The processed dataset\footnote{http://instruct-pix2pix.eecs.berkeley.edu/} consists of 454,445 examples, each example contains a text and an edited one. 

\paragraph{Metrics.}
Following prior works~\cite{instructpix2pix, SDEdit}, we use four metrics to evaluate the quality of image editing:
\begin{itemize}[leftmargin=*]
\item \emph{Image-Image similarity}. Cosine similarity of the original and edited CLIP~\cite{radford2021learning} image embeddings, which reflects how much the edited image agrees with the input image.

\item \emph{Directional CLIP similarity}~\cite{gal2022stylegan}. It represents how much the change in text agrees with the change in the images.

\item \emph{Fréchet Inception Distance (FID)}~\cite{heusel2017gans}. A metric compares the distribution of generated images with the distribution of a set of real images. Due to the absence of standard images in the dataset we used, we regard the images generated by vanilla stable diffusion text-to-image model (prompt with edited text) as the standard image set.

\item \emph{CLIP Accuracy}~\cite{hessel2021clipscore}. The percentage
of instances where the edited text has a higher similarity
to the target image, as measured by CLIP~\cite{radford2021learning}, than to the original source image, indicating how much the target image agrees with the edited text.

\end{itemize}

\subsection{Main Results}

\paragraph{Image Quality.}

\begin{figure*}[!h]
  \centering
  \includegraphics[width=\linewidth]{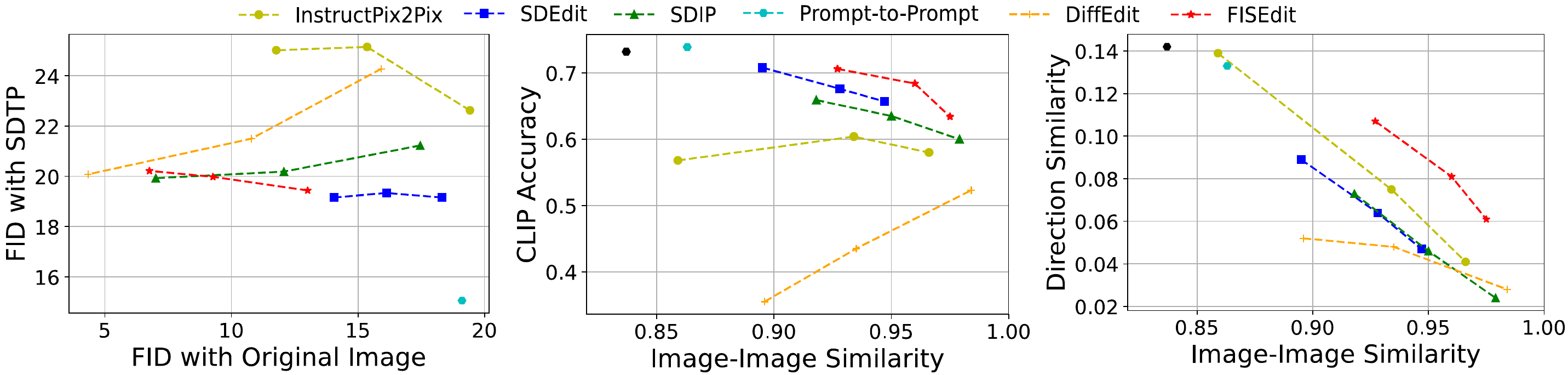}
  \caption{Quantitative results of the images generated by baselines and our method, demonstrating the trade-off between consistency with the input image and consistency with the semantic change. We vary InstructPix2Pix's image guidance scale between $[1.0, 2.5]$, SDEdit’s strength between $[0.5, 0.75]$, DIFFEdit’s strength between $[0.5, 1.0]$, and edited size between $[0.25, 0.75]$ for SDIP and our method.}
  \label{quantitative-result}
\end{figure*}

\begin{figure*}[!h]
  \centering
  \includegraphics[width=0.75\linewidth]{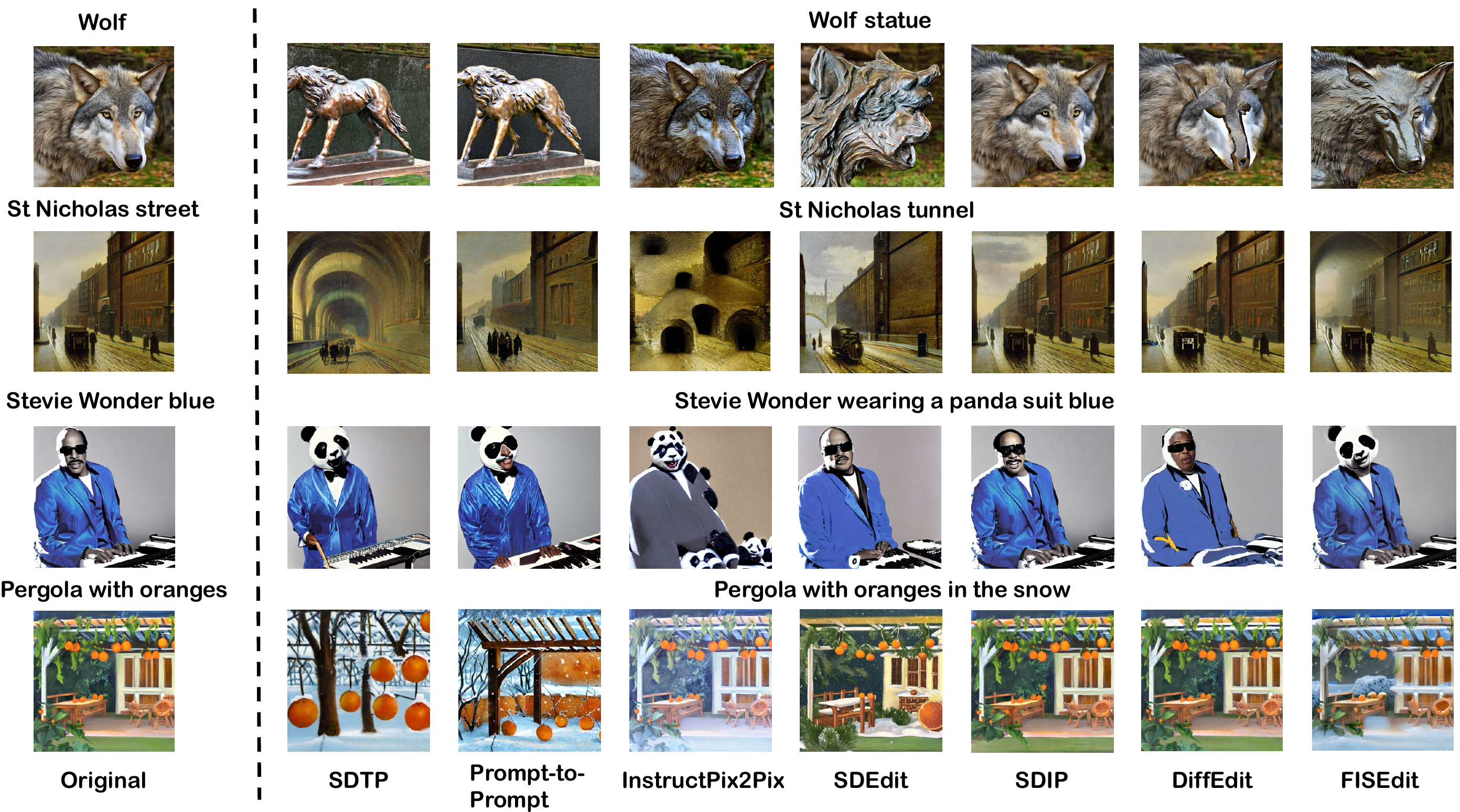}
  \caption{Qualitative results of the (768, 768) images generated by baselines and our method.}
  \label{qualitative-result}
\end{figure*}

Our main target in image quality is to achieve equal effect as baselines since we have reduced the computational complexity. We show the quantitative results in Figure~\ref{quantitative-result} and the qualitative results in Figure~\ref{qualitative-result}. We find that image-image similarity is competing with CLIP accuracy and directional CLIP similarity, which 
means that increasing the degree to which the output images correspond to a desired text will reduce their consistency with the input image. In order to show the tradeoff between two metrics, we have varied representative parameters for the methods. It can be inferred from these metrics that our method can better locate and modify changing areas as well as reflect the meaning of the text. 

\paragraph{Model Efficiency.}

We present efficiency comparison
results in Table~\ref{Model-efficiency-results}. We observe that all baselines struggle to generate or edit images efficiently since they cannot leverage the sparsity to avoid unnecessary computations. Worse, InstructPix2Pix~\cite{instructpix2pix} even encounter both computation and memory bottlenecks.
In contrast, our method can accurately edit the details of image, and also reduces the computation of diffusion models by up to $4.9\times$ when the image size is $768\times768$ and the edit size is 5\%. 
Eventually, we accelerate text-to-image inference by up to $4.4\times$ on NVIDIA TITAN RTX and $3.4\times$ on NVIDIA A100 when the edit size is 5\%.

\subsection{Ablation Study}
Table~\ref{ablation-op} shows the results of ablation studies conducted on TITAN RTX, validating the effectiveness of each improvement, and table~\ref{ablation-cache} outlines the impact of the cache system on data storage and transmission.

\begin{table}[ht]
\centering
\resizebox{\linewidth}{!}{%
    \begin{tabular}{ccccccc}
    \toprule
    \multirow{3}{*}{Size} & \multicolumn{4}{c}{Optimizations} & \multicolumn{2}{c}{Efficiency} \\ \cmidrule(r){2-5} \cmidrule(r){6-7}
     & \multirow{2}{*}{Op} & \multicolumn{2}{c}{Conv} & \multirow{2}{*}{Attn} & \multirow{2}{*}{Value} & \multirow{2}{*}{Ratio} \\ \cline{3-4}
     &  & w/o APSC & APSC &  &  &  \\ \hline
    \multirow{5}{*}{(512, 512)} &  &  &  &  & 5.42it/s & $1.0\times$ \\ \cline{2-7} 
     & \checkmark &  &  &  & 6.80it/s & $1.2\times$ \\ \cline{2-7} 
     & \checkmark & \checkmark &  &  & 6.86it/s & $1.2\times$ \\ \cline{2-7} 
     & \checkmark &  & \checkmark &  & 7.2it/s & $1.3\times$ \\ \cline{2-7} 
     & \checkmark &  & \checkmark & \checkmark & \textbf{10.17it/s} & \textbf{1.9}$\times$ \\ \hline
    \multirow{5}{*}{(768, 768)} &  &  &  &  & 1.83it/s & $1.0\times$ \\ \cline{2-7} 
     & \checkmark &  &  &  & 2.73it/s & $1.5\times$ \\ \cline{2-7} 
     & \checkmark & \checkmark &  &  & 3.17it/s & $1.7\times$ \\ \cline{2-7} 
     & \checkmark &  & \checkmark &  & 3.61it/s & $2.0\times$ \\ \cline{2-7} 
     & \checkmark &  & \checkmark & \checkmark & \textbf{8.13it/s} & \textbf{4.4}$\times$ \\ \hline
    \end{tabular}%
    }
    \caption{Ablation study of each optimization. Op: CUDA kernel improvements (e.g., Flash Attention~\cite{dao2022flashattention}). Conv: Combining tiled sparse convolution with or without APSC. Attn: Using approximate attention.}
    \label{ablation-op}
\end{table}

\begin{table}[ht]
  \centering
  \resizebox{\linewidth}{!}{
  \begin{tabular}{ccccccc}
    \toprule
    \multirow{3}{*}{Device} &  \multirow{3}{*}{Edit size}  & \multirow{3}{*}{Compute} & \multicolumn{2}{c}{Cached data (MB)} & \multicolumn{2}{c}{Transfer (ms)} \\  \cmidrule(lr){4-5} \cmidrule(lr){6-7}
    & & & w/o  & FISEdit & w/o  & FISEdit \\
    \midrule
    \multirow{4}*{TITAN RTX} & 30\% & 247ms & \multirow{4}*{4192} & 987 & \multirow{4}{*}{387} & 108 \\
     \cmidrule(lr){2-3} \cmidrule(lr){5-5} \cmidrule(lr){7-7} & 15\%  & 186ms &  & 1192 & & 114 \\
     \cmidrule(lr){2-3} \cmidrule(lr){5-5} \cmidrule(lr){7-7} & 5\%  & 128ms &  & 1348 & & 121 \\
    \midrule
    \multirow{4}*{A100 PCIe} & 30\% & 108ms & \multirow{4}*{4192} & 987 & \multirow{4}{*}{232} & 67 \\
     \cmidrule(lr){2-3} \cmidrule(lr){5-5} \cmidrule(lr){7-7} & 15\% &  84ms & & 1192 &  & 72 \\
     \cmidrule(lr){2-3} \cmidrule(lr){5-5} \cmidrule(lr){7-7} & 5\% & 72ms & & 1348 &  & 76 \\
    \bottomrule
  \end{tabular}
  }
  \caption{Comparison of computation time, data transfer time and cached data size with and without FISEdit.}
  \label{ablation-cache}
\end{table}

\begin{table*}[!t]
  \centering
    \resizebox{0.75\linewidth}{!}{%
    \begin{tabular}{@{}ccccccccccc@{}}
    \toprule
    \multirow{3}{*}{Method} & \multicolumn{2}{c}{Property} & \multicolumn{7}{c}{Efficiency} \\ \cmidrule(l){2-3} \cmidrule(l){4-10} 
     & \multirow{2}{*}{No Extra Training} & \multirow{2}{*}{Partially Edit} & \multirow{2}{*}{Edit Size} & \multicolumn{2}{c}{MACs} & \multicolumn{2}{c}{TITAN RTX} & \multicolumn{2}{c}{A100 PCle 40GB} \\ \cmidrule(l){5-6} \cmidrule(l){7-8} \cmidrule(l){9-10} 
     &  &  &  & Value & Ratio & Value & Ratio & Value & Ratio \\ \midrule
    SDTP & \checkmark & $\times$ & \multirow{3}{*}{-} & \multirow{3}{*}{1901G} & \multirow{3}{*}{$1.0\times$} & \multirow{3}{*}{1.83it/s} & \multirow{3}{*}{$1.0\times$} & \multirow{3}{*}{3.86it/s} & \multirow{3}{*}{$1.0\times$} \\ 
    SDIP & $\times$ & \checkmark &  &  &  &  &  &  &  \\ 
    Prompt-to-Prompt & \checkmark & $\times$ &  &  &  &  &  &  &  \\ \midrule
    pix2pix-zero & \checkmark & $\times$ & \multirow{2}{*}{-} & \multirow{2}{*}{-} & \multirow{2}{*}{-} & \multirow{2}{*}{OOM} & \multirow{2}{*}{-} & \multirow{2}{*}{OOM} & \multirow{2}{*}{-} \\
    PPAP & \checkmark & $\times$ &  & & & & & & \\
    \midrule
    InstructPix2Pix & $\times$ & $\times$ & - & 4165G & $0.5\times$ & OOM & - & 2.29it/s & $0.6\times$ \\ \midrule
    SDEdit & \checkmark  & $\times$ & - & 1521G & $1.2\times$ & 2.28it/s & $1.2\times$ & 4.82it/s & $1.2\times$ \\ \midrule
    DIFFEdit & \checkmark & \checkmark & - & 3041G & $0.6\times$ & 1.15it/s & $0.6\times$ & 2.42it/s & $0.6\times$ \\ \midrule
    \multirow{4}{*}{FISEdit} & \multirow{4}{*}{\checkmark} & \multirow{4}{*}{\checkmark} & 
     30\% & 835G & $2.3\times$ & 4.06it/s & $2.2\times$ & 9.57it/s & $2.5\times$ \\ \cmidrule(l){4-10} 
     &  &  & 15\% & 485G & $3.9\times$ & 5.64it/s & $3.1\times$ & 11.6it/s & $3.0\times$ \\ \cmidrule(l){4-10} 
     &  &  & 5\% & \textbf{386G} & \textbf{4.9}$\times$ & \textbf{8.14it/s} & \textbf{4.4}$\times$ & \textbf{13.1it/s} & \textbf{3.4}$\times$ \\ \bottomrule
    \end{tabular}%
}
\caption{Property and efficiency evaluation of each method. We use the Multiply-Accumulate Operations (MACs) of U-Net to measure computational cost and the number of U-Net calls completed per second as a measure of speed, as U-Net calls are the computational bottleneck of the above models. Due to the fact that the inference speed of the baseline models will not be affected by the size of modifications, we don't discuss edit size for them. The strength of SDEdit and DIFFEdit are set to 0.8.}
\label{Model-efficiency-results}
\end{table*}

\section{Conclusion}
In real-world text-to-image model applications, users often desire minor edits to generated images based on textual descriptions. However, existing methods fail to exploit the redundant computation between previous and current diffusion model calls. To sufficiently reuse prior results, we have solved the three challenges: difference detection, sparse computation and pre-computed data management. The experimental results indicate that our method can accurately edit the details of images while achieves a considerable speedup.

\paragraph{Limitations.}
Our method has a poor performance when editing a low-resolution images (e.g., $256\times256$). This is because, the image changes caused by semantic modification do not exhibit adequate sparsity in low-resolution images. However, GANs hold a competitive position in low-resolution image generation, and have less computational overhead compared to diffusion models, meaning that large image generation is the main challenge for real-world deployment.

\paragraph{Prospects.} 
With our method, it is possible to design a cache system for a real-world text-to-image service. Specifically, when a user query that contains a text is received, we can search for the most similar text-image pairs stored in the cache and process this query sparsely and incrementally, thereby achieving better system throughput.

\section*{Acknowledgements}
This work is supported by National Natural Science Foundation of China (U23B2048 and U22B2037), China National Postdoctoral Program for Innovative Talents (BX20230012), and PKU-Tencent joint research Lab. Bin Cui and Xupeng Miao are the co-corresponding authors.

\bibliography{main}
\appendix

\setcounter{table}{0}
\setcounter{figure}{0}
\renewcommand{\thetable}{\arabic{table}}
\renewcommand{\thefigure}{\arabic{figure}}

\onecolumn

\noindent \huge{\sc \textbf{Appendix}} \\
\LARGE{\sc \textbf{Additional Results}}

\begin{figure*}[!h]
  \centering
  \includegraphics[width=\linewidth]{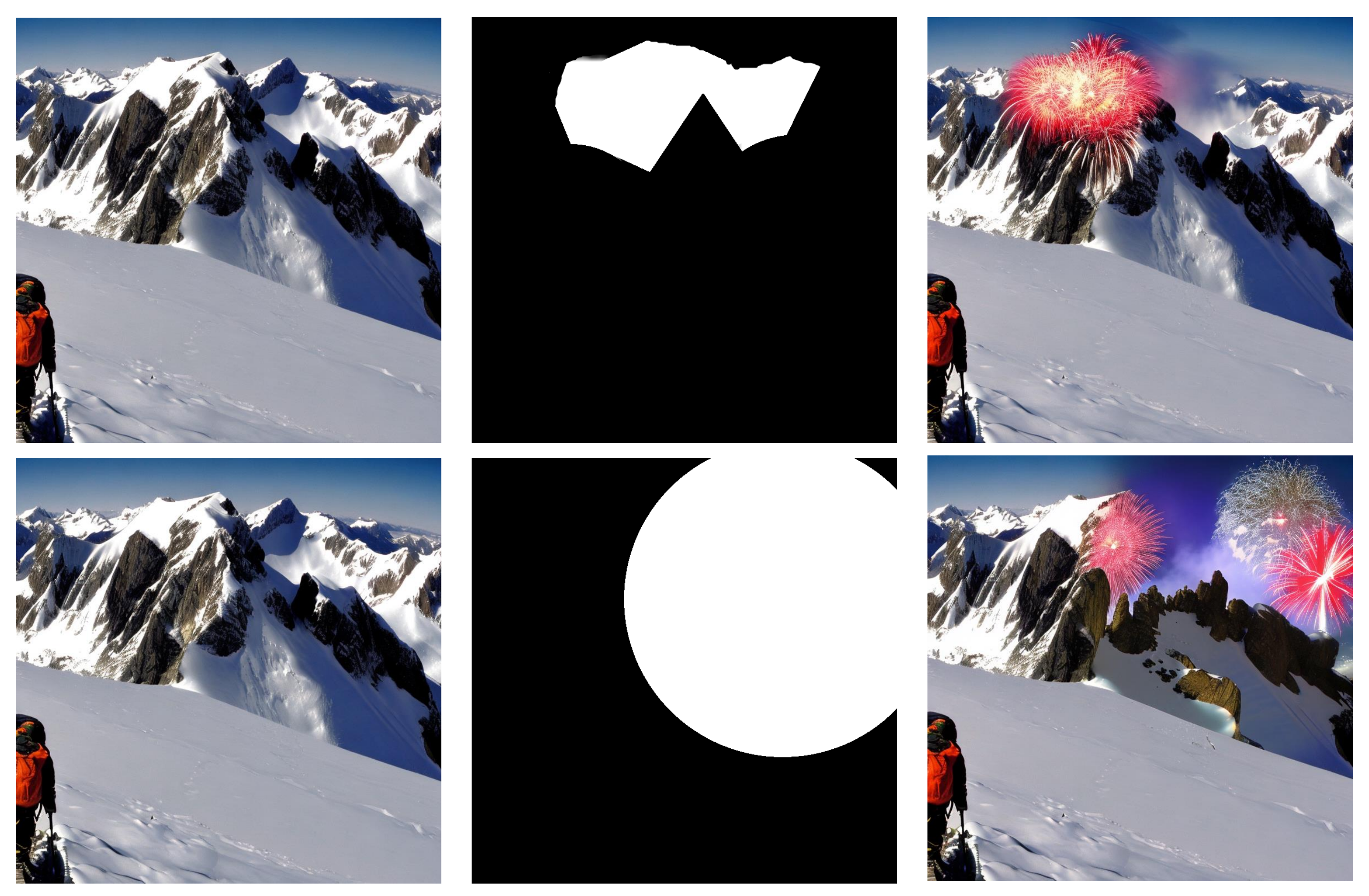}
  \caption{
  In real-world scenarios, users may desire the ability to provide masks and selectively modify only the regions within the mask that they consider unsatisfactory.
  As shown in the provided example, 
  The user specifies different masks and edits prompt from \emph{"Mountaineering Wallpapers"} to \emph{"Mountaineering Wallpapers under fireworks"}.
  }
\end{figure*}

\begin{figure*}[!h]
  \centering
  \includegraphics[width=0.9\linewidth]{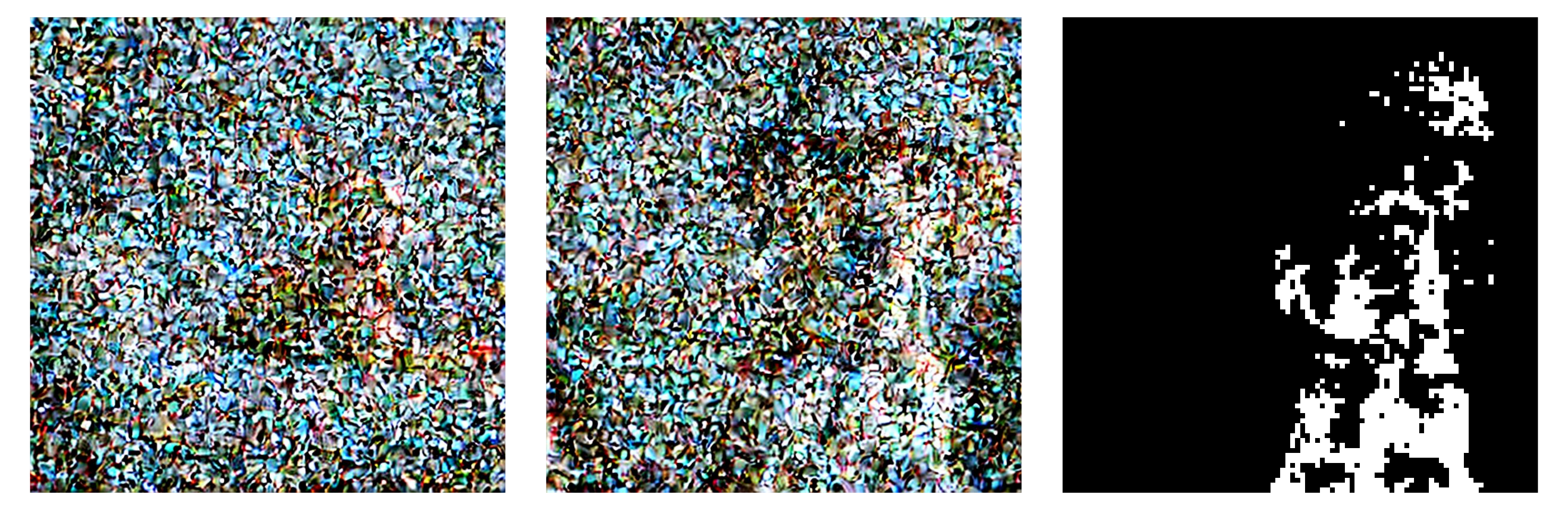}
  \includegraphics[width=0.9\linewidth]{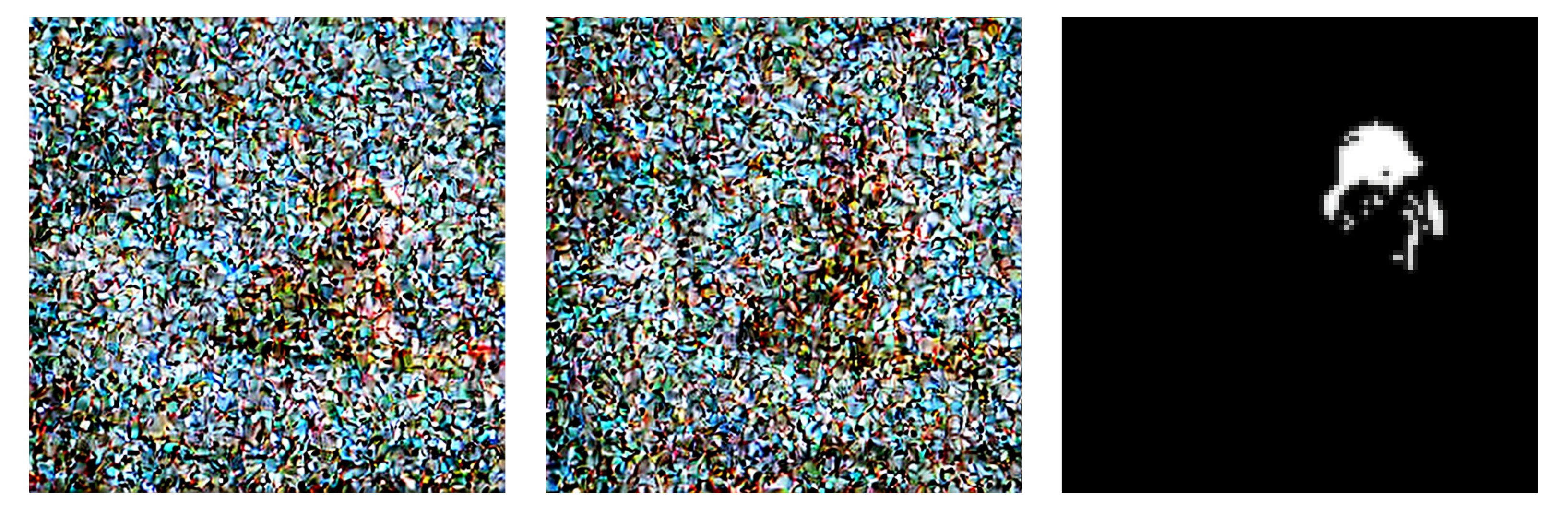}
  \caption{Visualization of masks generated with (lower) and without (upper) cross-attention control. For both, the corresponding prompts are \emph{“A dog is sitting on the sofa”} and \emph{“A dog is sitting on the sofa with a hat on its head”}.}
\end{figure*}

\begin{figure*}[t]
  \centering
  \includegraphics[width=1.0\linewidth]{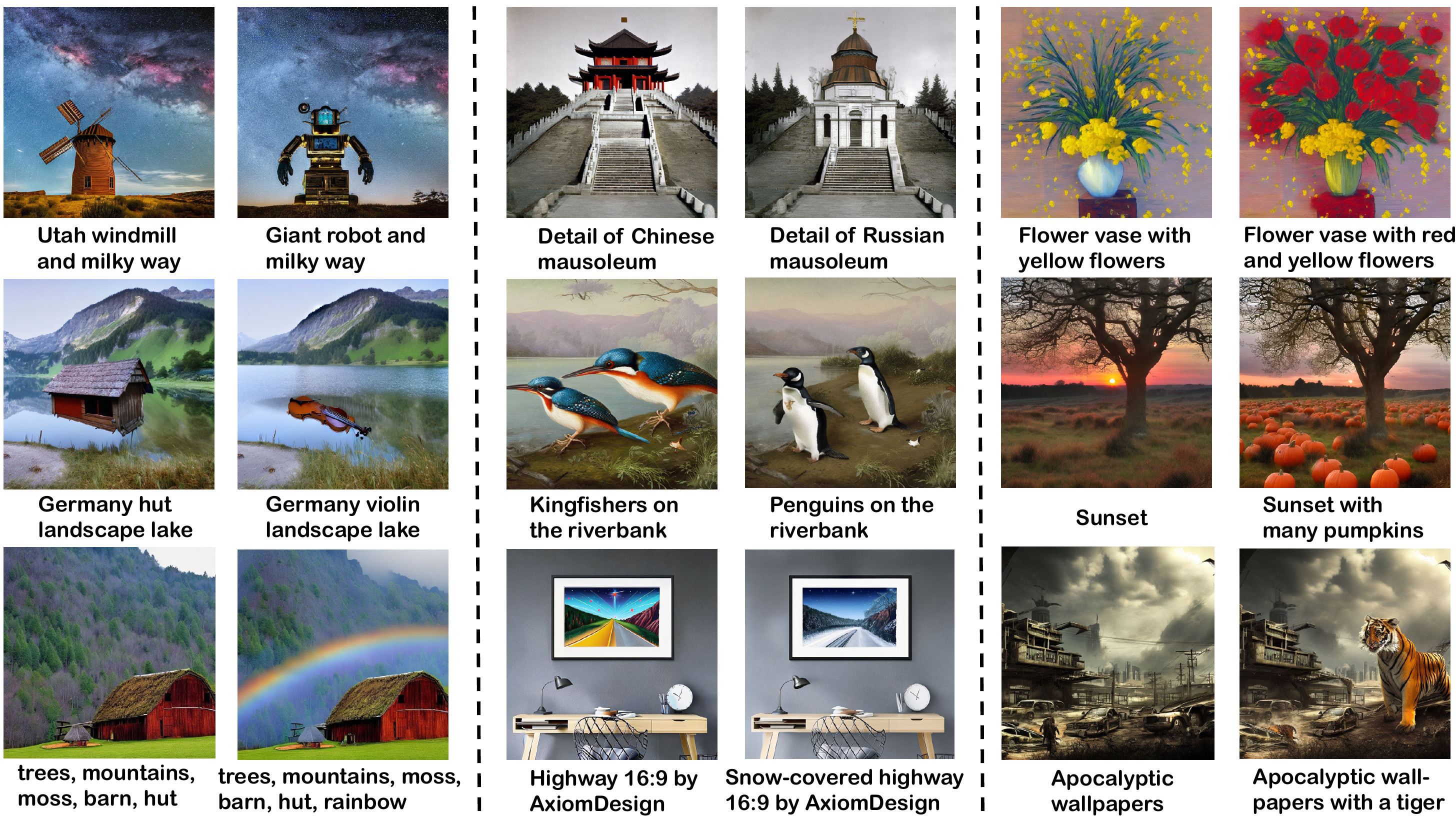}
  \caption{Additional results regarding the images generated by FISEdit, indicating 
  its ability to handle more complex situations.}
\end{figure*}

\end{document}